\RequirePackage[T1]{fontenc}
\documentclass[letterpaper,10pt,conference]{ieeeconf}

\IEEEoverridecommandlockouts
\usepackage{amsmath}
\usepackage{amssymb}
\usepackage{balance}
\usepackage{booktabs}
\usepackage{cite}
\usepackage{graphicx}
\usepackage{placeins}
\usepackage{microtype}
\usepackage{xcolor}
\usepackage{xspace}
\usepackage{url}
\usepackage[hidelinks]{hyperref}

\graphicspath{{figs/}}

\newcommand{\R}{\mathbb{R}}
\newcommand{\loss}{\mathcal{L}}

\newcommand{\GRiD}{GRiD\xspace}

\title{\LARGE\bf
cuRoboV2-Fast: Fast Torque-Limited Collision-Free Motion Planning
}

\author{%
\authorblockN{Balakumar Sundaralingam\qquad
Adithyavairavan Murali\qquad Stan Birchfield\\[3pt]}
\authorblockA{\includegraphics[width=20mm]{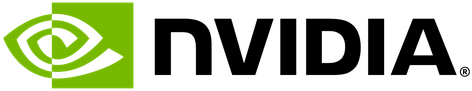}}}

\makeatletter
\IEEEaftertitletext{%
  \begin{minipage}{\textwidth}
    \centering
    \includegraphics[width=\textwidth]{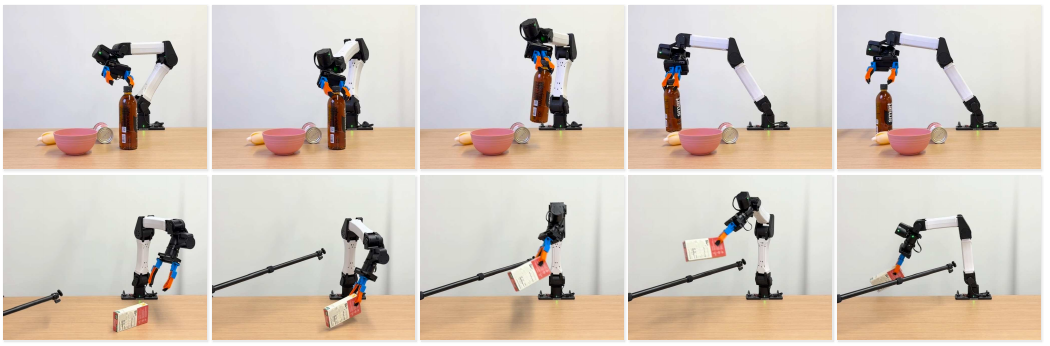}
    \def\@captype{figure}%
    \caption{Collision-free, torque-limited manipulation with cuRoboV2-Fast:
    payload transfer (top) and manipulation with a moving obstacle (bottom).
    Our stereo-camera-based deployment achieves 5.93\,s mean pick-and-drop
    cycles, 34.6\% faster cycle time than cuRoboV2.}
    \label{fig:hero}
  \end{minipage}
  \vspace{0.6em}
}
\makeatother

\begin{document}

\raggedbottom

\maketitle
\thispagestyle{empty}
\pagestyle{empty}

\begin{abstract}
We present cuRoboV2-Fast, a GPU-accelerated planner that reduces both planning
latency and execution time for collision-free, payload-aware, torque-limited
motions. B-spline trajectory optimization reduces the required iterations,
while accelerated robot computations and differentiable inverse dynamics reduce
refinement cost and accommodate changing payloads. TimeOpt optimizes traversal
speed along the fixed path under sampled velocity, acceleration, jerk, and torque
constraints, exploiting local spline support to solve banded Newton systems
efficiently on the GPU. With a 3\,kg payload, cuRoboV2-Fast achieves
4.4\,ms mean planning time on RTX 5090, $14.7\times$ faster than cuRoboV2. Our method
on Jetson AGX Orin plans faster than cuRoboV1 + TOPPRA
on RTX 5090. A stereo-camera-based
deployment on YAM achieves 5.93\,s mean completed pick-and-drop cycles,
34.6\% shorter than cuRoboV2.
Code (Apache 2.0): \href{https://github.com/NVlabs/curobo}{\nolinkurl{github.com/NVlabs/curobo}}.
Project Page: \href{https://research.nvidia.com/labs/lpr/curobov2-fast/}{\nolinkurl{research.nvidia.com/labs/lpr/curobov2-fast/}}.

\par
\end{abstract}
\begin{figure*}[!t]
 \centering
 \includegraphics[width=\textwidth]{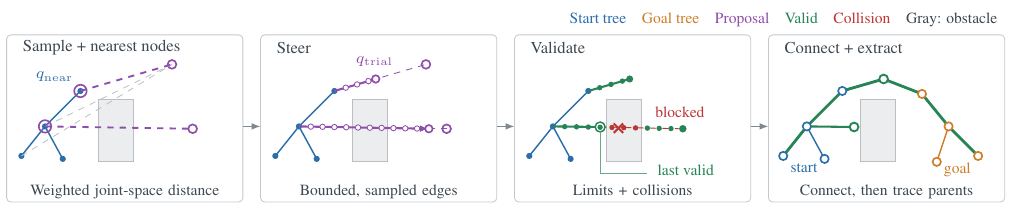}
 \caption{Our RRT evaluates multiple tree extensions in parallel, validates their edges separately, and connects the start and goal trees to recover a route. A parallel scan inserts accepted endpoints, retaining valid prefixes of blocked edges. Purple segments show unchecked proposals, while green segments have passed validation.}
 \label{fig:collision_free_rrt}
\end{figure*}

\section{Introduction}
\label{sec:introduction}

Fast manipulation requires motions that are both quick to plan and quick to
execute. Motion planning has made substantial progress in finding collision-free
paths through cluttered environments~\cite{thomason2024vamp}. Executing these
paths requires timing that respects velocity, acceleration, and actuator torque
limits. Carried payloads change the dynamics and can render an otherwise feasible
motion infeasible. Moreover, geometric path length alone does not determine
execution time: a shorter path can require slower motion under actuator
constraints. Efficient motion planning must therefore consider both the
computational cost of generating a trajectory and the time required to execute it.

Prior work establishes the benefits of parallel planning, smooth trajectory
representations, and efficient robot
computations~\cite{curobo_icra23,pan2011collision,plancher2022grid}.
Planning executable motions nevertheless requires repeated collision and dynamics
evaluations during trajectory optimization. Reducing this cost requires both
fewer optimization iterations and cheaper computations within each iteration.
Execution time presents a complementary challenge: uniformly rescaling the
timestep cannot redistribute time along a path whose limiting actuator
constraints vary with configuration. Efficient payload-aware planning therefore
requires flexible optimization of trajectory geometry and timing alongside
fast constraint evaluation.

We present cuRoboV2-Fast, a single-GPU planner for collision-free, torque-limited
motions that reduces both planning latency and execution time. B-spline
trajectory optimization refines geometry with fewer iterations, while TimeOpt
optimizes traversal speed along the resulting fixed path. The timing spline's
local support yields banded systems for efficient GPU optimization. A parallel
GPU RRT provides route initialization when needed, and accelerated robot
computations reduce the cost of refinement. Differentiable inverse dynamics
incorporates payload-dependent torque limits and supports runtime inertial
updates as carried objects change.
We make six contributions toward fast, executable motion planning:
\begin{enumerate}
    \item We use B-splines for collision-free, payload-aware, torque-limited
    trajectory optimization, matching the success of cuRoboV1~\cite{curobo_icra23}
    with 25 versus 125 initial iterations in our convergence benchmark.
    \item We introduce GPU-accelerated TimeOpt with banded Newton systems and
    a primal--dual predictor--corrector solver for fixed-path timing under
    sampled velocity, acceleration, jerk, and torque bounds. It reduces mean
    motion duration by 15.7\% beyond uniform timestep refinement and averages
    0.28\,ms on valid 100\,g benchmark queries on RTX 5090.
    \item We accelerate robot computations and differentiable inverse dynamics,
    supporting payload updates without recompilation. Trajectory optimization iterations are $2.7\times$ faster than cuRoboV2~\cite{curobo_v2}
    on RTX 5090 and $2.6\times$ faster on Jetson AGX Orin.
    \item We achieve sub-millisecond route initialization with a parallel
    GPU RRT: 0.305\,ms on average, $43\times$ faster than
    cuRoboV1 PRM~\cite{curobo_icra23} and $1.48\times$ faster than
    pRRTC~\cite{huang2026prrtc} under coarse collision checks.
    Separating constraint validation from tree growth enables additional
    checks without changing growth kernels.
    \item Our integrated pipeline achieves 4.4\,ms mean motion-planning time
    with a 3\,kg payload on RTX 5090, $14.7\times$ faster than
    cuRoboV2~\cite{curobo_v2} while maintaining similar success.
    cuRoboV2-Fast on Jetson AGX Orin also plans faster than
    cuRoboV1~\cite{curobo_icra23} +
    TOPPRA~\cite{pham2018toppra} on RTX 5090.
    \item We demonstrate a stereo-camera-based pick-and-drop pipeline on YAM,
    integrating perception, grasp generation, and motion planning on a single
    GPU. Mean completed cycle time is 5.93\,s, 34.6\% shorter than cuRoboV2,
    with planning accounting for only 1.0\% of cycle time.
\end{enumerate}

We have open sourced the cuRoboV2-Fast implementation under the Apache 2.0
license at \href{https://github.com/NVlabs/curobo}{\nolinkurl{github.com/NVlabs/curobo}}.

\section{Related Work}
\label{sec:related_work}

Recent advances in motion planning exploit parallelism at several levels.
Building on RRT-Connect~\cite{kuffner2000rrt}, VAMP uses CPU SIMD instructions
to accelerate kinematics and collision checking~\cite{thomason2024vamp}, while
pRRTC distributes tree expansion, nearest-neighbor search, and edge validation
across GPU threads~\cite{huang2026prrtc}. cuRoboV1 combines geometric initialization
with parallel trajectory optimization to generate collision-free
motions~\cite{curobo_icra23}. These systems demonstrate the benefits of designing
planning operations around the available hardware. Our RRT follows this direction
while keeping validation separate from tree growth, so additional checks can be
introduced without modifying the growth kernels.

Once an initial path is available, trajectory optimization can improve its
smoothness and clearance, as demonstrated by CHOMP, STOMP, and
TrajOpt~\cite{ratliff2009chomp,kalakrishnan2011stomp,Schulman2014MotionPW}.
The trajectory representation also shapes this refinement: B-splines have long
supported collision-free smoothing and reactive deformation, using local
support to modify portions of a path~\cite{pan2011collision,chiaravalli2018bspline}.
Earlier whole-body optimization further shows how collision and torque
constraints can be considered together~\cite{lengagne2013generation}.
Building on these ideas, we study how parallel spline evaluation and gradient
accumulation make constrained refinement efficient within a complete planner.

Fixed-path timing methods optimize traversal speed while preserving geometry.
Bobrow et al.~\cite{bobrow1985timeoptimal} consider actuator torque bounds,
and Verscheure et al.~\cite{verscheure2009timeoptimal} formulate path timing
as convex optimization. TOPPRA~\cite{pham2018toppra} handles velocity,
acceleration, and torque constraints through backward reachability and forward
construction using small linear programs. Its path-restricted torque is affine
in path acceleration and squared speed, allowing dynamics coefficients to be
precomputed and reused. TOPP3~\cite{pham2017topp3} addresses third-order
constraints by integrating maximum and minimum jerk profiles and connecting
them through multiple shooting.

Our TimeOpt in cuRoboV2-Fast adopts the same fixed-path dynamics reduction and directly optimizes
a cubic B-spline representation of squared speed under sampled velocity,
acceleration, jerk, and torque bounds. Each sample depends on four timing
controls, yielding banded Newton systems for a primal--dual predictor--corrector
solver. This local support and parallel constraint evaluation enable efficient
GPU optimization, with a mean TimeOpt latency of 0.28\,ms across valid 100\,g
benchmark queries on RTX 5090.

Trajectory optimization repeatedly evaluates robot kinematics and collisions and
propagates their derivatives. Efficient rigid-body algorithms and kinematic
optimization libraries, including Pinocchio and PyRoki, provide reusable
building blocks for this work~\cite{carpentier2019pinocchio,kim2025pyroki}.
For collisions, broad-phase pruning reduces the geometry requiring detailed
tests~\cite{pan2012fcl}, while differentiable GPU evaluation supports parallel
trajectory optimization in cuRoboV1~\cite{curobo_icra23}. We build on these
principles with shared kinematic gradient accumulation, parallel Jacobians,
and pruning of self-collision and swept world queries. Their use in both
trajectory refinement and grasp selection motivates evaluation at batch size
one as well as larger batches.

Torque-aware optimization adds repeated dynamics and derivative evaluations to
this workload. \GRiD accelerates these operations through robot-specific code
generation, sparsity, and GPU parallelism~\cite{plancher2022grid}.
We focus on directly computing gradients of the optimization loss and supplying
chain schedules and inertial parameters as runtime data. This avoids forming
full dynamics derivative matrices and permits payload updates without
recompilation when the reduced topology is unchanged, while retaining
specialization for robot dimensions.
Integrating these methods with spline optimization and geometric initialization enables collision-free, torque-limited motion and grasp planning on a single GPU.

\section{Approach}
\label{sec:approach}

Our motion planner combines collision-free initialization, B-spline
optimization, TimeOpt for trajectory retiming, efficient robot computations, and
differentiable inverse dynamics. Planning first refines a straight-line seed and invokes RRT when
needed. Constrained motions, such as grasp approach and lift, use cost terms
that heavily penalize deviations from the prescribed path.

\subsection{Collision-Free Initialization with RRT}
\label{sec:collision_free_initialization}

When a straight-line seed is insufficient, we initialize optimization with a
route from a bounded bidirectional RRT~\cite{kuffner2000rrt}. The planner grows
trees from the start and goal in alternating waves, evaluating multiple
extensions in parallel. Each extension samples a configuration, finds its
nearest tree node, and steers toward the sample with a bounded step before
checking joint limits and self/world collision (Fig.~\ref{fig:collision_free_rrt}).
Accepted endpoints extend the active tree, including the valid prefix of an
otherwise blocked edge, and the opposite tree attempts to connect to them.
Once the trees meet, tracing their parent chains recovers the route. The search
stops when a connection is found or a preset limit on stored nodes or expansion
rounds is reached.

Our implementation organizes growth into synchronized rounds of parallel
extensions. In contrast, pRRTC~\cite{huang2026prrtc} lets GPU blocks repeatedly
sample and grow trees asynchronously, with greedy connection steps inside each
block. Our nearest-node kernel distributes distance calculations across both
stored nodes and joint coordinates, processing the tree in tiles and reducing
the results to select each candidate's nearest node. In pRRTC, each thread
computes complete distances for a subset of nodes before a block-wide minimum
reduction. Thus, both planners parallelize nearest-node search, but distribute
its work differently.

We check sampled states along all proposed edges together, then determine each
edge's valid prefix. A parallel scan assigns accepted endpoints distinct storage
slots, allowing concurrent insertion without reserving each slot atomically as
in pRRTC. Each round also evaluates a bounded extension from the opposite tree.
This organization keeps validation separate from growth, allowing additional
checks without changing the growth kernels. Reusing allocated buffers and
replaying growth operations through CUDA graphs reduces repeated dispatch
overhead. Once a route is found, we shorten it and fit a spline seed for
refinement.

\subsection{B-Spline Trajectory Optimization}
\label{sec:bspline}

\begin{figure}[t]
 \centering
 \includegraphics[width=\columnwidth]{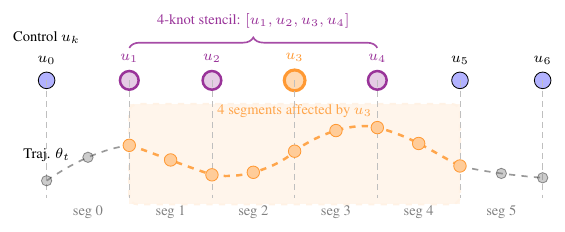}
 \caption{Spline evaluation and backward gather: each sample uses four controls; $u_3$ gathers weighted sample gradients from the four orange segments and sums them using warp-level reduction primitives.}
 \label{fig:bspline_local_support}
\end{figure}

\paragraph{Parallel evaluation.}
We use the local support of uniform cubic B-splines to parallelize trajectory
evaluation and its backward pass. Let $U\in\R^{K\times d}$
contain the free controls for $d$ joints, $h$ be the knot interval, and
$\alpha\in[0,1]$ the coordinate within segment $k$:
\begin{equation}
 q^{(r)}(t)=h^{-r}\sum_{j=0}^{3}N_j^{(r)}(\alpha)\,\widetilde u_{k+j},
 \qquad r=0,1,2,3,
 \label{eq:spline_evaluation}
\end{equation}
Here $N_j$ are cubic basis functions and $\widetilde U$ is the full control
sequence. This established
representation~\cite{pan2011collision,chiaravalli2018bspline} gives each sample
four supporting controls (Fig.~\ref{fig:bspline_local_support}). We assign each
sample--joint pair to a GPU thread, which evaluates position, velocity,
acceleration, and jerk together from those controls. Samples therefore execute
independently, with adjacent threads processing adjacent joints for coalesced
memory access. Increasing the sampling density adds parallel evaluation work
without increasing the number of optimized controls.

\paragraph{Gradient accumulation.}
The backward pass reverses this mapping: each control gathers gradients only
from samples within its support. In Fig.~\ref{fig:bspline_local_support},
$u_3$ gathers the sample gradients from the four orange segments, weights them
by the corresponding basis derivatives, and sums them into its control gradient.
Samples outside this region do not contribute. Writing sampled derivatives as
$q_i^{(r)}=A_i^{(r)}U+b_i^{(r)}$, we accumulate
\begin{equation}
 \nabla_U\loss=\sum_i\sum_{r=0}^{3}
       (A_i^{(r)})^{\mathsf T}\,\nabla_{q_i^{(r)}}\loss.
 \label{eq:spline_gradient}
\end{equation}
The kernel applies the basis weights directly rather than materializing the
matrices $A_i^{(r)}$. Threads assigned to a control--joint pair split the sample
contributions across warp lanes. Each lane combines the incoming position,
velocity, acceleration, and jerk gradients with their corresponding basis
weights; a warp reduction then sums the partial contributions. One lane writes
the resulting control gradient. This gather avoids atomic accumulation from
overlapping samples and a sample-by-control intermediate tensor, while allowing
different controls and joints to be processed independently.

\subsection{TimeOpt: Fixed-Path Timing Optimization}
\label{sec:trajectory_retiming}

In Sec.~\ref{sec:bspline}, we optimize position controls at a fixed knot
interval $\Delta t$ and reduce this interval between passes. This uniform
reduction is limited by the most constrained part of the trajectory.
TimeOpt optimizes timing along the fixed path, allowing other portions
to be traversed faster while respecting actuator limits.

Let $q(s)$ be the fixed position spline, where $s\in[0,1]$ denotes progress
along the path. A second cubic B-spline represents squared path speed, with
timing controls $r=(r_0,\ldots,r_{L-1})\in\R^L$ as the optimization variables:
\begin{equation}
 x_r(s)=\dot{s}^{\,2}=\sum_{\ell=0}^{L-1}B_\ell(s)r_\ell.
 \label{eq:retiming_speed}
\end{equation}
Here $B_\ell(s)$ is a clamped cubic B-spline basis function and $r_\ell$
is its squared-speed coefficient. Both splines are functions of path progress
$s$. Positive coefficients give $x_r(s)>0$; forward traversal follows
$\dot{s}=\sqrt{x_r(s)}$, where the dot denotes a time derivative.

TimeOpt seeks the minimum-duration traversal of this fixed path subject
to velocity, acceleration, jerk, and torque limits:
\begin{equation}
 \begin{gathered}
 \min_r\quad \widehat T(r)=\frac{1}{N+1}\sum_{i=0}^{N}\frac{1}{\sqrt{x_r(s_i)}}\\
 \begin{aligned}
 \text{subject to}\quad
 & r_\ell>0, && \ell=0,\ldots,L-1,\\
 & |\dot q_r(s_i)|\leq\bar v, && \forall i,\\
 & |\ddot q_r(s_i)|\leq\bar a, && \forall i,\\
 & |\dddot q_r(s_i)|\leq\bar j, && \forall i,\\
 & \tau_{\min}\leq\tau_r(s_i)\leq\tau_{\max}, && \forall i.
 \end{aligned}
 \end{gathered}
 \label{eq:retiming_problem}
\end{equation}
The objective averages inverse path speed at fixed uniform samples $s_i=i/N$
to approximate duration. The bounds $\bar v$, $\bar a$, $\bar j$, and
$[\tau_{\min},\tau_{\max}]$ limit joint velocity, acceleration, jerk, and
torque, respectively. We also enforce these limits at timing knots and on both
sides of internal path knots, and preserve the prescribed initial velocity
and acceleration.

\begin{figure}[t]
 \centering
 \includegraphics[width=\columnwidth]{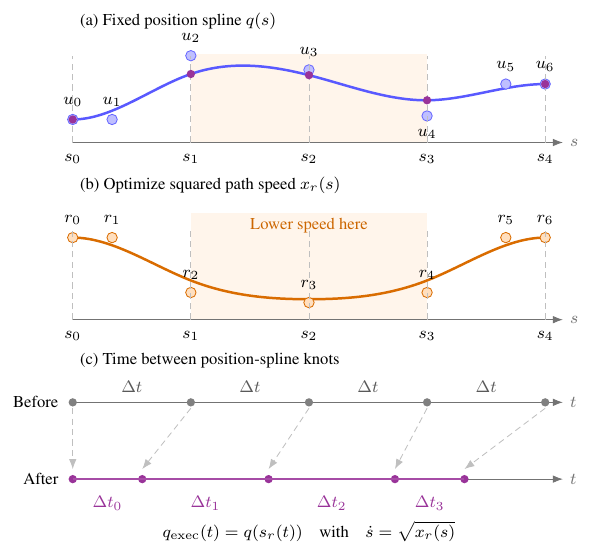}
 \caption{TimeOpt schematic. (a) The position spline and its control
 points $u_i$ remain fixed; purple markers indicate positions at path knots.
 (b) Optimizing coefficients $r_i$ changes the squared path speed $x_r(s)$.
 (c) The resulting time map assigns unequal traversal times to the same path
 intervals. In this example, slower traversal through the shaded region and
 faster traversal elsewhere reduce the total duration.}
 \label{fig:retiming_time_map}
\end{figure}

The timing spline determines when the robot reaches each point on the path:
\begin{equation}
 t_r(s)=\int_0^s\frac{1}{\sqrt{x_r(u)}}\,du,
 \qquad q_{\mathrm{exec}}(t)=q(s_r(t)).
 \label{eq:retiming_composition}
\end{equation}
Here $s_r(t)$ is the inverse of $t_r(s)$. Thus, TimeOpt changes when each path
point is reached while preserving the path (Fig.~\ref{fig:retiming_time_map}).

We use the standard fixed-path dynamics
reduction~\cite{verscheure2009timeoptimal,pham2018toppra} to evaluate dynamics
once and reuse them as timing changes. Torque depends on squared path speed
through the exact affine relation
\begin{equation}
 \tau_r(s)=\tfrac12 A(s)x_r'(s)+D(s)x_r(s)+g(s),
 \label{eq:retiming_torque}
\end{equation}
where primes denote path derivatives. The vectors $A(s)$ and $D(s)$ capture
inertial effects, and $g(s)$ is gravity torque. Using the inverse-dynamics
computation in Sec.~\ref{sec:inverse_dynamics}, we obtain them from three
evaluations per path sample: $(q,0,0)$, $(q,0,q')$, and
$(q,q',q'')$, yielding $g$, $A+g$, and $D+g$, respectively, as in the TOPPRA
implementation~\cite{toppra_joint_torque}.
Subsequent timing updates use Eq.~\ref{eq:retiming_torque} without repeating
inverse dynamics. Joint velocity, acceleration, and jerk follow from the
chain rule using the fixed path derivatives.

\begin{figure}[t]
 \centering
 \includegraphics[width=\columnwidth]{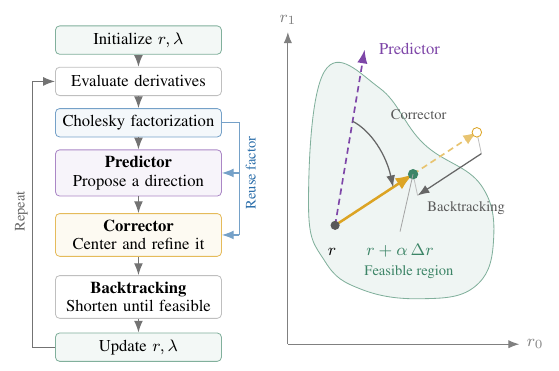}
 \caption{TimeOpt solver. The predictor and corrector reuse
 one Cholesky factorization. In the schematic, the corrector changes the
 purple direction to yellow; backtracking shortens the step to the green
 feasible update.}
 \label{fig:retiming_solver}
\end{figure}

We solve Eq.~\ref{eq:retiming_problem} with the primal--dual
predictor--corrector method illustrated in Fig.~\ref{fig:retiming_solver}.
The solver updates timing controls $r$ and positive constraint multipliers
$\lambda$. For each inequality, a positive margin $m_j$ measures the distance
to its bound. The predictor targets zero complementarity, $m_j\lambda_j=0$;
the corrector balances these products across constraints to keep the search
in the interior and accounts for their second-order change.
The timing spline's local support yields banded Newton systems, and sampled
constraints are evaluated in parallel on the GPU. Both direction calculations
reuse the same Cholesky factorization. Backtracking selects a step satisfying
the nonlinear bounds, and we retain the shortest feasible iterate. The final
trajectory undergoes sampled actuator and collision checks and verification
of its endpoint conditions.

TimeOpt targets efficient convergence to a local minimum and does not guarantee
global optimality. It reduces mean motion duration by over 15\% relative to
trajectory refinement with uniform timestep rescaling
(Sec.~\ref{sec:trajopt_convergence}). These motion-time gains and low
computational cost make TimeOpt useful for real-robot deployment.

\subsection{Accelerated Robot Computations}
\label{sec:gpu_computation}

We reduce the cost of repeated robot evaluations by preparing topology-dependent
data once and sharing intermediate results across computations. GPU tensor
operations identify each joint's descendants and the joints contributing to
each tool--actuator pair. Using this structure, forward kinematics computes
local link poses in parallel, then composes world poses in tree order through
shared memory. The resulting transforms provide the tool poses and collision
spheres and support Jacobian evaluation.

\begin{figure}[t]
 \centering
 \includegraphics[width=\columnwidth]{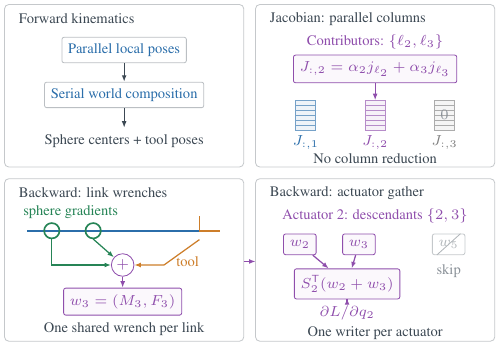}
 \caption{Efficient kinematics: parallel local poses feed serial world composition, while prepared contributors enable independent Jacobian columns (top). The backward pass accumulates link wrenches once, then gathers them per actuator, avoiding repeated ancestor walks (bottom).}
 \label{fig:kinematics_combined}
\end{figure}

For the backward pass, we first combine output gradients into one wrench per
link. Each actuator independently gathers the wrenches of affected links,
projects them onto its joint motion directions, and combines coupled-joint
contributions before writing its gradient. Accumulating the link wrenches once
avoids repeatedly walking the ancestor chain for every output. When IK requires
an explicit Jacobian, we instead assign its columns to independent workers,
each using the prepared contributor lists; trajectory optimization consumes
the loss gradients directly (Fig.~\ref{fig:kinematics_combined}). These fast
Jacobians support Levenberg--Marquardt initialization of the IK solver. In the figure,
$S_2$ denotes a world-frame joint screw and $\alpha_\ell$ describes
actuator-to-joint coupling.

\begin{figure}[t]
 \centering
 \includegraphics[width=\columnwidth]{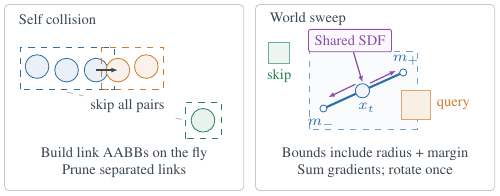}
 \caption{Collision pruning and reuse. On-the-fly link AABBs eliminate sphere--sphere tests between separated links (left). Sweep bounds prune distant obstacles; adjacent half-sweeps share the SDF at $x_t$ and rotate their summed gradient once (right). $m_\pm$ denote neighboring midpoints.}
 \label{fig:self_collision_aabb}
 \label{fig:world_sweep}
 \label{fig:collision_execution}
\end{figure}

\begin{figure}[t]
 \centering
 \includegraphics[width=\columnwidth]{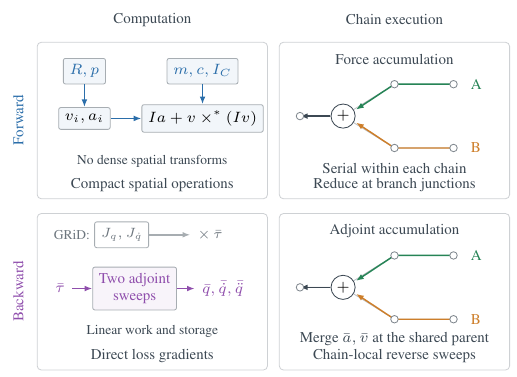}
 \caption{Efficient inverse dynamics. Forward RNEA uses compact spatial operations (top); backward computes loss gradients directly instead of forming full derivative matrices (bottom). Independent chains run concurrently, reducing at shared parents. Trees show leaf-to-root accumulation. Payload updates require no recompilation.}
 \label{fig:rnea_schedule}
\end{figure}

Self-collision checking tests for overlap between collision spheres on the
robot's links. cuRoboV1~\cite{curobo_icra23} evaluates configured sphere--sphere
pairs in parallel, but the number of checks can become costly for robots
represented by many spheres. We compute an axis-aligned bounding box (AABB) for each link on the
fly from its current sphere positions and padded radii. These bounds are built
in parallel across links and stored in shared memory within the collision
kernel. When two link AABBs do not overlap, we skip all sphere--sphere checks
between those links. This inexpensive pruning can substantially reduce detailed
pair tests, particularly when links are well separated. Each GPU block combines
the remaining tests, cost reduction, and sphere gradients for one robot state.

Swept world collision uses the same principle to prune distant cuboids. Bounds
that include the sphere radius, activation margin, and adjacent half-segments
first reject distant cuboids. The remaining half-sweeps share the distance
evaluation at their common trajectory point and combine their gradients before
a single rotation to world coordinates (Fig.~\ref{fig:collision_execution}).
This reduces repeated geometry calculations while retaining the chosen
discrete sampling resolution for validation.

\begin{figure*}[t]
\centering
\includegraphics[width=\textwidth]{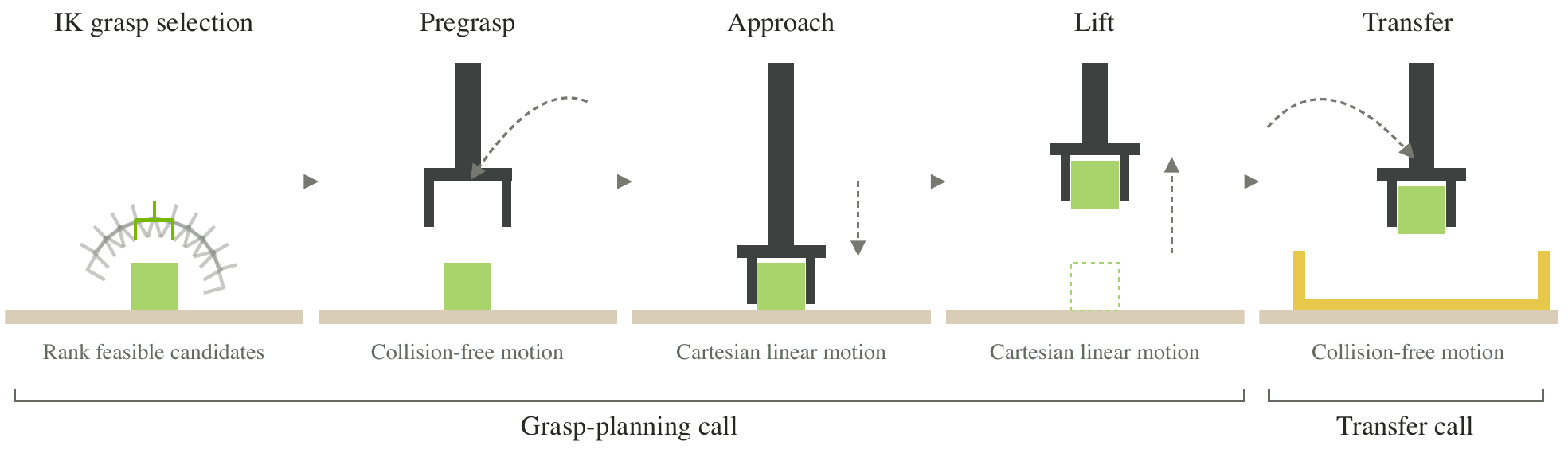}
\caption{Grasp-planning integration. IK selects a grasp, followed by pregrasp,
approach, and lift motions in one planning call. Approach and lift follow
Cartesian linear motions. A separate call plans transfer to the placement pose.}
\label{fig:grasp_planning_integration}
\end{figure*}

\subsection{Differentiable Inverse Dynamics and Payloads}
\label{sec:inverse_dynamics}

\GRiD exploits robot-specific code generation to optimize dynamics and its
derivatives~\cite{plancher2022grid}. Our implementation instead keeps the chain
schedule and inertial parameters as data, while specializing kernel dimensions
and execution options at compile time. Efficiency comes from compact spatial
operations, chain-local execution, and direct loss-gradient computation;
payload updates do not require regenerating or recompiling the kernels.

At setup, we merge fixed bodies and prepare the chain schedule on the CPU.
Independent chains execute concurrently, with each GPU worker traversing a
serial chain locally. This avoids synchronization between successive joints
within a chain; workers synchronize only at inter-chain dependencies.
Spatial transforms use rotations and translations directly rather than dense
$6\times6$ transform matrices. The
recursive Newton--Euler algorithm computes link velocities, accelerations,
and forces as
\begin{align}
 v_i &= X_i v_{\pi(i)}+S_i\dot q_i,\nonumber\\
 a_i &= X_i a_{\pi(i)}+S_i\ddot q_i+v_i\times S_i\dot q_i,\nonumber\\
 f_i &= I_i a_i+v_i\times^*(I_iv_i)-f_i^{\mathrm{ext}},
 \label{eq:rnea_forward}
\end{align}
where $X_i$ is a spatial transform, $S_i$ the joint motion subspace, and $I_i$
the spatial inertia. Forces are then accumulated from leaves toward the root,
and projection onto each joint's motion subspace gives
$\tau_i=S_i^{\mathsf T}f_i$. Payload changes update mass, center of mass, and
inertia at runtime; the schedule remains valid while the reduced topology is
unchanged.

Our analytical backward pass maps torque-loss gradients directly to
$(q,\dot q,\ddot q)$ and then to spline controls through
Eq.~\ref{eq:spline_gradient}. This avoids forming and contracting the full
configuration and velocity derivative matrices returned by
\GRiD~\cite{plancher2022grid}, reducing both derivative computation and output
storage. Both passes combine sibling contributions with warp reductions and
reuse shared-memory storage after its values are consumed or saved. The
backward pass also reuses forward intermediates to avoid recomputation.
Figure~\ref{fig:rnea_schedule} summarizes the computation and chain execution.
These savings preserve runtime updates to the
robot's inertial parameters.

\subsection{Grasp-Planning Integration}
\label{sec:grasp_pipeline}

An object can be grasped from many positions and orientations, so grasp planning
begins by choosing which pose to plan for. Given a set of candidate grasp poses,
we use IK to find feasible robot configurations and rank them by pose error and
distance from the current configuration. We found this sufficient for grasp
selection in our evaluated task, allowing us to replace trajectory optimization
over the grasp-pose goal set with a less expensive IK solve. The selected pose
defines the target for subsequent motion planning; IK feasibility alone does
not establish that the robot can reach it along a collision-free path.

We then plan three connected motions (Fig.~\ref{fig:grasp_planning_integration}): a move to a pregrasp pose offset from the
selected grasp, an approach along the grasp axis, and a lift of the object.
Spline optimization generates the pregrasp motion, with RRT initialization when
needed to navigate obstacles, while constrained spline optimization produces
Cartesian linear motions for approach and lift. Each motion
starts from the preceding motion's terminal state, with collision settings
adjusted to permit the intended contact.
A separate planning call generates a collision-free transfer of the grasped
object to the placement pose.

\FloatBarrier
\section{Experiments}
\label{sec:experiments}
We evaluate route initialization, torque-limited trajectory optimization,
robot computations, and pick-and-place planning. Motion benchmarks use
Franka Panda (7-DoF), computation benchmarks also include G1 (29-DoF),
and manipulation tasks use YAM. Offline planning runs on RTX 5090, with
additional computation timings on Jetson; real-robot deployment uses RTX 6000 Ada.

\subsection{Coarsely Checked Route Initialization}
\begin{figure}[!htbp]
\centering
\includegraphics[width=\columnwidth]{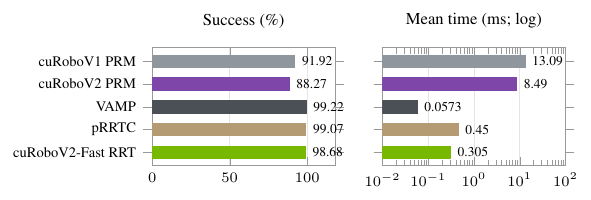}
\caption{Coarsely checked route initialization on MotionBenchMaker and
MPINets (2,575 queries). Our RRT implementation is $1.48\times$
faster than pRRTC. VAMP plans routes faster, but VAMP + TOPPRA
achieves lower motion-planning success with payloads than cuRoboV2-Fast
(Fig.~\ref{fig:full_motion}).}
\label{fig:graph_planners}
\end{figure}

\begin{figure}[!tp]
\centering
\includegraphics[width=\columnwidth]{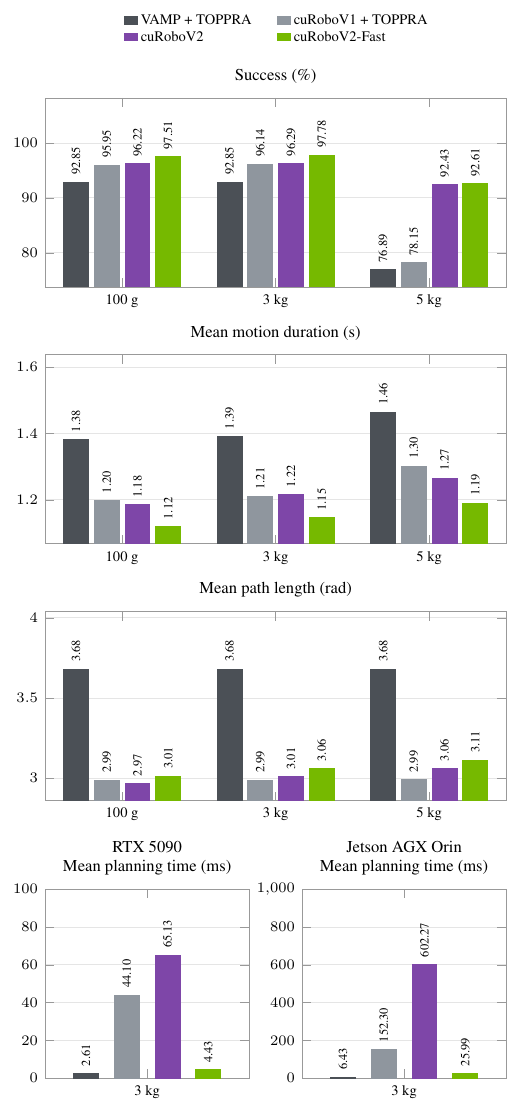}
\caption{Motion Planning with different payloads. cuRoboV2-Fast achieves
similar success to cuRoboV2 while planning faster and producing shorter
motions. Both V2 methods retain higher success than the TOPPRA baselines
at the heaviest payload. Quality panels use RTX 5090 results; Fast success
averages two repeats of 2,590 queries per payload. Motion duration and path
length use 1,599 queries valid for all methods and payloads, including both
Fast repeats. Separate planning-time panels show the 3\,kg payload on
RTX 5090 and Jetson AGX Orin, using different timing scales.}
\label{fig:full_motion}
\end{figure}

We evaluate routes used to initialize trajectory optimization, with coarse
collision checks at up to 0.03\,rad per-joint spacing.
On 2,575 queries from MotionBenchMaker~\cite{chamzas2022motionbenchmaker}
and MPINets~\cite{fishman2022mpinets}, our RRT implementation
achieves 98.68\% initialization success, compared with 91.92\% for
cuRoboV1 PRM~\cite{curobo_icra23} and 88.27\% for cuRoboV2 PRM
(Fig.~\ref{fig:graph_planners}). Mean planning time is 0.305\,ms,
versus 13.09\,ms and 8.49\,ms for the two PRM baselines, corresponding
to $43.0\times$ and $27.9\times$ speedups. Our RRT is also $1.48\times$
faster than pRRTC~\cite{huang2026prrtc}, which achieves 99.07\% success at 0.450\,ms.
VAMP~\cite{thomason2024vamp} achieves 99.22\% success at 0.0573\,ms; although it initializes
routes faster, VAMP + TOPPRA~\cite{pham2018toppra} has lower motion-planning success with
payloads than our pipeline (Fig.~\ref{fig:full_motion}).

Timing and path length use 2,101 shared-success queries. Planning time
includes native shortcutting and excludes IK. Mean path length is
4.93\,rad for ours, compared with 5.28\,rad for cuRoboV1 PRM,
3.70\,rad for cuRoboV2 PRM, 5.00\,rad for VAMP, and 5.32\,rad for pRRTC.
Routes that fail the coarse collision checks are rejected. Accepted
routes serve as optimization seeds and require further refinement and
validation.

\subsection{Trajectory Optimization Convergence}
\label{sec:trajopt_convergence}
\begin{figure}[!htbp]
\centering
\includegraphics[width=\columnwidth]{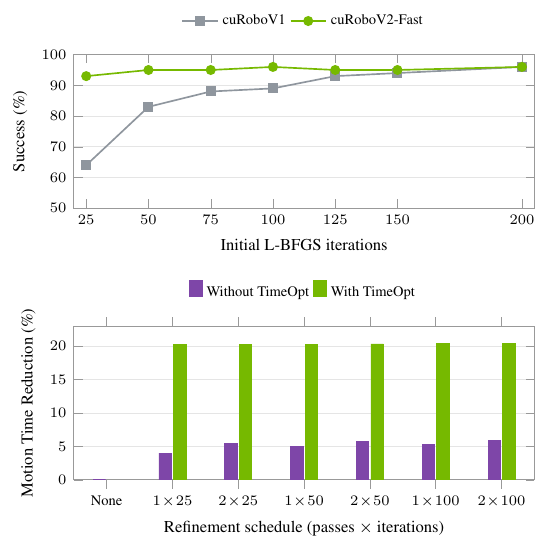}
\caption{Trajectory optimization convergence on 100 MotionBenchMaker
\texttt{cage\_panda} queries. Top: initial-only success.
Bottom: cuRoboV2-Fast mean motion-time reduction on 93 shared valid queries
after 25 initial iterations. Our B-spline formulation achieves higher
success at small initial budgets. TimeOpt reduces mean motion time by
16\% compared with finetuning using fixed timestep rescaling.}
\label{fig:trajopt_convergence}
\end{figure}

We isolate the effects of initial optimization, timestep refinement,
and TimeOpt on 100 MotionBenchMaker \texttt{cage\_panda} queries.
cuRoboV1~\cite{curobo_icra23} and cuRoboV2-Fast use the same
four saved IK seeds per query and a single attempt, with dynamics and
particle optimization disabled. Figure~\ref{fig:trajopt_convergence}
shows that our B-spline formulation achieves higher success at small
iteration budgets: cuRoboV2-Fast reaches 93\% validated success with
25 initial L-BFGS iterations, compared with 64\% for cuRoboV1.
cuRoboV1 requires $5\times$ as many initial iterations to match this
success rate. Increasing our initial budget to 200 raises success by only
three percentage points.

We then fix our initial budget at 25 iterations and compare one or two
refinement passes of 25, 50, or 100 iterations, using a timestep-shrink
factor of 0.8. All settings retain 93\% success.
On the 93 queries valid across all settings, two 25-iteration passes
reduce mean motion duration by 5.5\% relative to initial optimization
alone. Adding TimeOpt increases the total reduction to 20.3\%,
corresponding to a further 15.7\% reduction relative to the refined
trajectory. Longer refinement schedules improve motion duration after TimeOpt by
less than 0.2\%.

For the reported cuRoboV2-Fast timings and real-robot deployment, we use
50 initial iterations followed by one 25-iteration refinement pass
($50 + 1\times25$), giving a total budget of 75 iterations.

\subsection{Motion Planning with Payloads}
\label{sec:payload_results}
We study the full pipeline for generating a collision-free, torque-limited
motion for execution, given a start state and a target pose.
For methods that do not consider torque limits during planning, we
postprocess their trajectories using TOPPRA~\cite{pham2018toppra}, which
changes timing along a fixed path.
We compare four pipelines: cuRoboV2-Fast, cuRoboV2~\cite{curobo_v2},
cuRoboV1~\cite{curobo_icra23} + TOPPRA, and
VAMP~\cite{thomason2024vamp} + TOPPRA.

As shown in Figure~\ref{fig:full_motion}, cuRoboV2-Fast and cuRoboV2
achieve similar success rates across payloads,
exceeding those of cuRoboV1 + TOPPRA and VAMP + TOPPRA.
At the heaviest payload, success decreases for all methods, with a larger
drop for the two TOPPRA baselines; cuRoboV2-Fast retains a
14.5--15.7 percentage-point advantage over them.

On queries valid for all methods at all payloads, cuRoboV2-Fast produces
the shortest mean motion duration, reducing duration by 5.8\%
relative to cuRoboV2 on average across payloads.
Path length alone can be misleading when comparing motion efficiency.
cuRoboV2-Fast produces mean joint-space paths that are 0.8--4.2\% longer
than those of cuRoboV1 + TOPPRA, yet its motions are 6.8\% shorter
in duration on average across payloads.

For the 3\,kg planning-time comparison, cuRoboV2-Fast takes 4.43\,ms
on RTX 5090, $14.7\times$ faster than cuRoboV2 and
$10.0\times$ faster than cuRoboV1 + TOPPRA.
VAMP + TOPPRA plans $1.7\times$ faster than cuRoboV2-Fast,
but achieves lower success
and excludes IK from its timing.
On Jetson AGX Orin, cuRoboV2-Fast takes 25.99\,ms, $23.2\times$ faster
than cuRoboV2 and $5.9\times$ faster than cuRoboV1 + TOPPRA
(Fig.~\ref{fig:full_motion}). Its relative speedup over cuRoboV2
is greater than on RTX 5090.
\textbf{cuRoboV2-Fast on Jetson AGX Orin plans faster than
cuRoboV1 + TOPPRA on RTX 5090.}

\subsection{Robot Computation Performance}
\begin{figure*}[!tp]
\centering
\includegraphics[width=\textwidth]{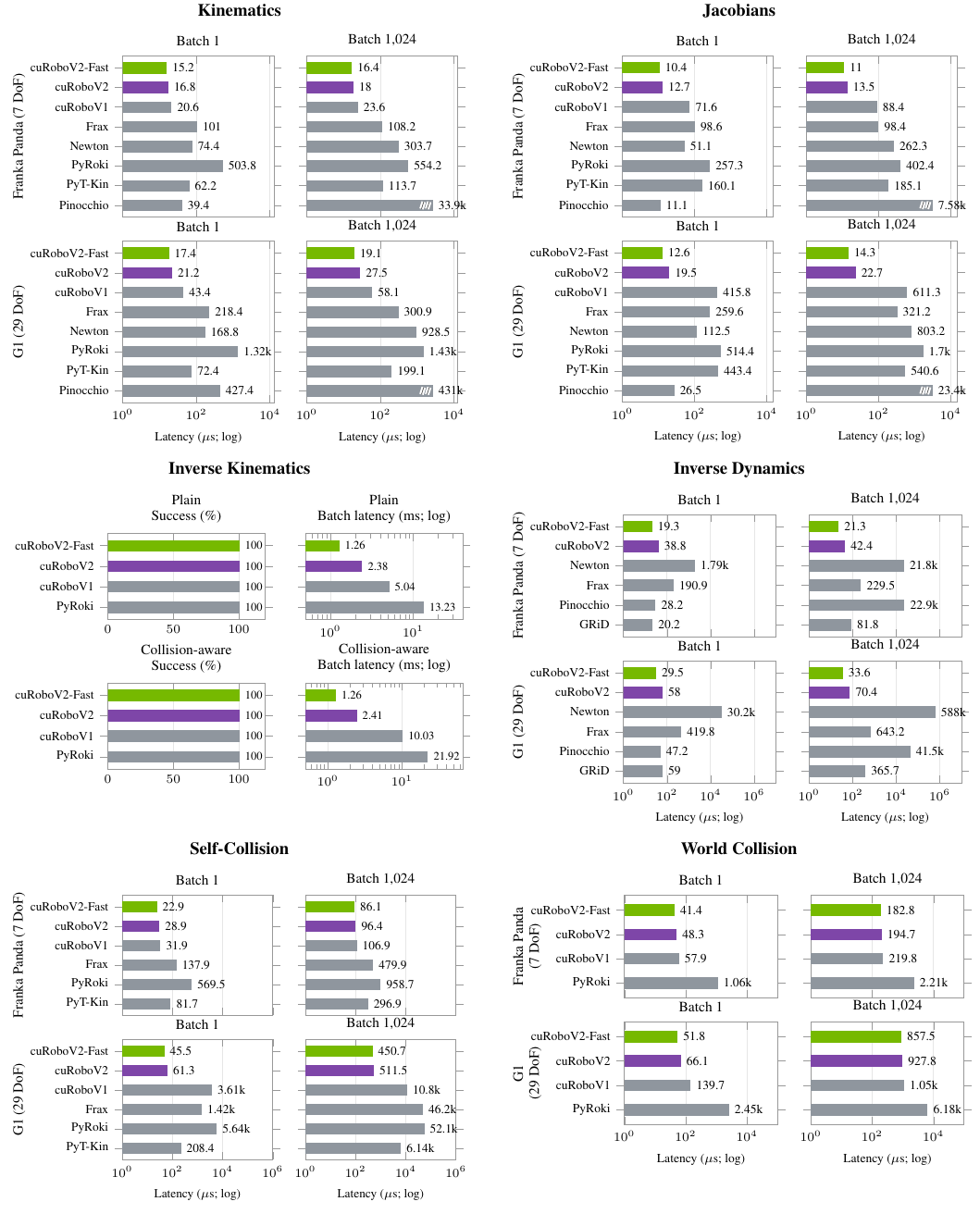}
\caption{Robot computation performance on RTX 5090. cuRoboV2-Fast
accelerates robot computations and collision-aware IK. Latencies are medians;
IK uses 100 shared Franka Panda targets. Other modules compare batches
1 and 1,024 on Franka Panda and G1. Broken-bar labels show full values;
k denotes $10^3$.}
\label{fig:robot_computation_grid}
\end{figure*}

Having established the planning-time speedups of the full motion planning
pipeline, we now examine computation times for individual modules.
We benchmark median latency on RTX 5090 for Franka Panda and G1 at batches
1 and 1,024, comparing our implementation with cuRoboV1, cuRoboV2, Frax, Newton,
PyRoki, PyT-Kin (pytorch\_kinematics), Pinocchio, and GRiD where applicable.
Module speedups average latency ratios over the two batch sizes;
comparisons covering both robots also average over Franka and G1.

Kinematics scales to large batches with little additional latency:
increasing the batch from 1 to 1,024 adds less than 10\% on either robot.
Our forward and backward computations are $1.1\times$ faster than
cuRoboV2 on Franka and $1.3\times$ faster on G1
(Fig.~\ref{fig:robot_computation_grid}).
Jacobian gains are larger on the higher-DoF G1 robot
(Fig.~\ref{fig:robot_computation_grid}). Our implementation is $1.2\times$
faster than cuRoboV2 on Franka and $1.6\times$ faster on G1.
On G1, the speedup over cuRoboV1 is approximately $38\times$.
Our IK solver uses explicit Jacobians during Levenberg--Marquardt
initialization. On 100 shared Franka Panda targets, all four IK methods
achieve 100\% success under common pose, joint-limit, and collision checks.
cuRoboV2-Fast is $1.9\times$ faster than cuRoboV2 for collision-aware IK,
preserving success while reducing batch latency
(Fig.~\ref{fig:robot_computation_grid}).

Inverse dynamics takes approximately half as long as cuRoboV2 while
supporting runtime changes to inertial parameters
(Fig.~\ref{fig:robot_computation_grid}). Increasing the batch from 1 to
1,024 adds less than 20\% latency on either robot.
Relative to GRiD~\cite{plancher2022grid}, the reported latency ratios average
$2.4\times$ on Franka and $6.4\times$ on G1 across the two batch sizes.
At batch 1, our GPU implementation is also faster than CPU Pinocchio
on both robots, with an average speedup of $1.5\times$.

Collision computations improve over cuRoboV2 at both batch sizes
(Fig.~\ref{fig:robot_computation_grid}),
with mean speedups of $1.22\times$ for self-collision and $1.15\times$
for swept world-collision. G1 self-collision shows a much larger gain
over cuRoboV1, with a speedup of approximately $52\times$.

\begin{figure*}[!tp]
\centering
\includegraphics[width=\textwidth]{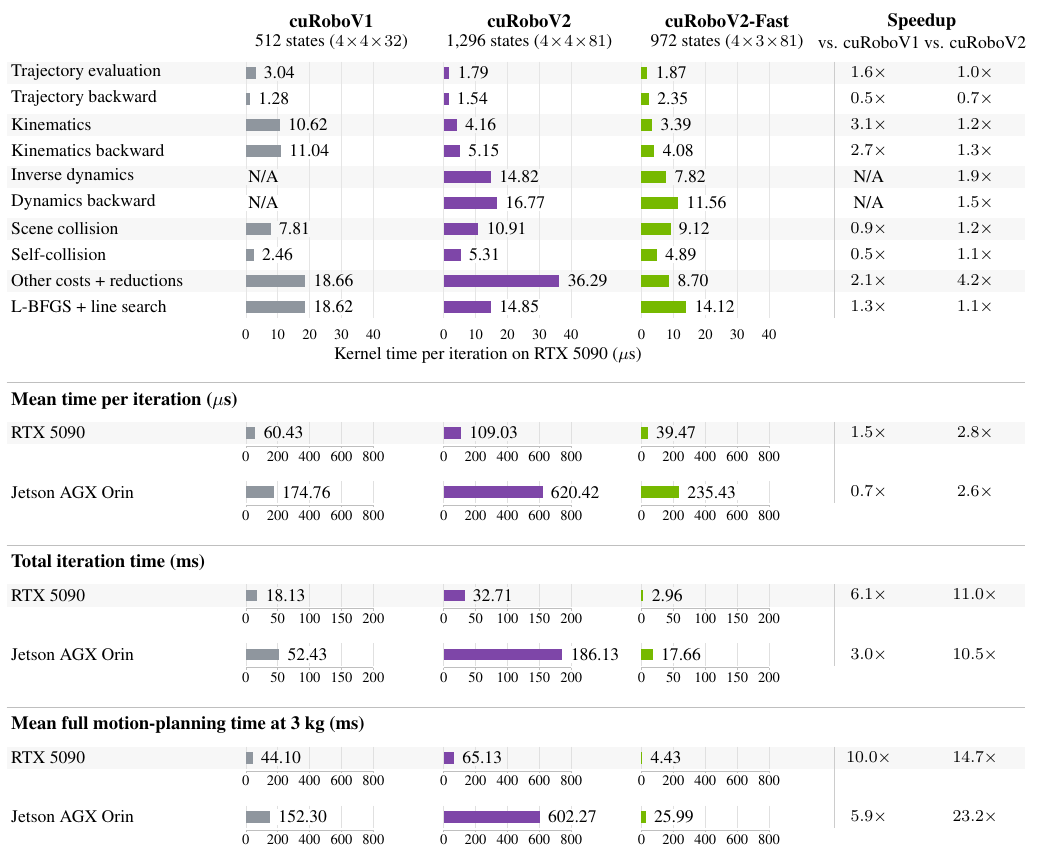}
\caption{Trajectory-optimization and full motion-planning times.
Faster iterations and a smaller iteration budget compound to reduce optimization time. cuRoboV2-Fast achieves 11.0$\times$ and 10.5$\times$ shorter optimization-loop times than cuRoboV2 on RTX 5090 and Jetson AGX Orin, respectively. Iteration totals use 75 iterations for cuRoboV2-Fast and 300 for both baselines.
Iteration timings on both devices use 100 repeated
\texttt{bookshelf\_small\_panda} solves at 100\,g.}
\label{fig:trajopt_module_timings}
\end{figure*}

\begin{figure}[!tp]
\centering
\includegraphics[width=\columnwidth]{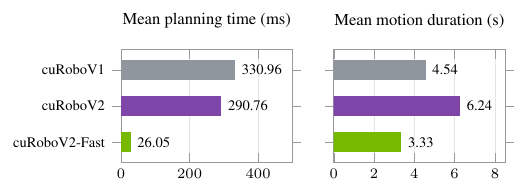}
\caption{Mean per-cube planning time and planned motion duration for the
five-cube simulated YAM tray arrangement. cuRoboV2-Fast plans over
$11\times$ faster and produces shorter motions than both baselines.}
\label{fig:yam_tray_comparison}
\end{figure}

\begin{figure}[!htbp]
\centering
\includegraphics[width=\columnwidth]{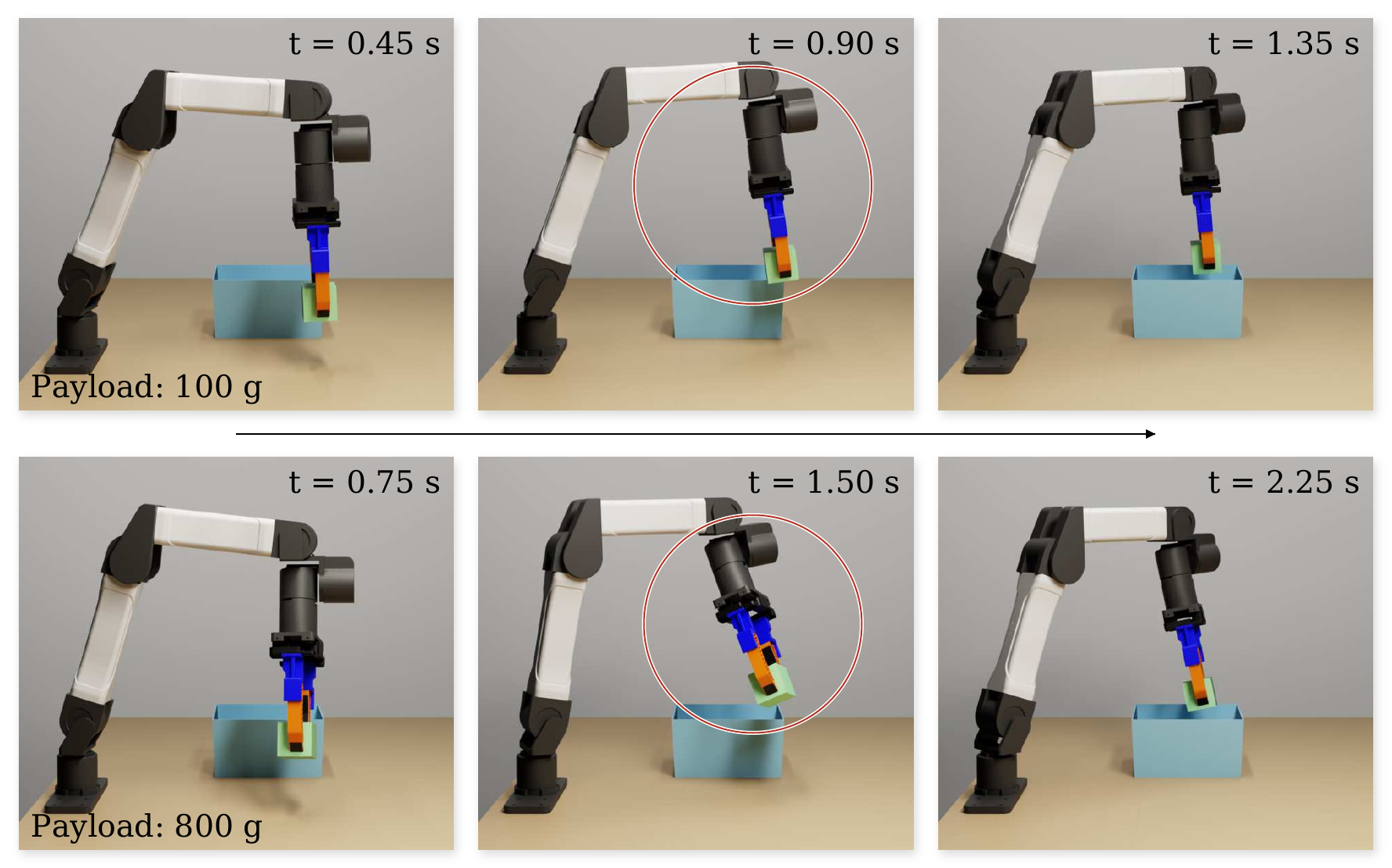}
\caption{Simulated YAM transfers with 100\,g (top) and 800\,g (bottom)
payloads. The heavier payload shifts motion from joint 2 to joint 5 to
respect joint 2's torque limit.}
\label{fig:yam_payload_comparison}
\end{figure}

\begin{figure*}[!tp]
\centering
\includegraphics[width=\textwidth]{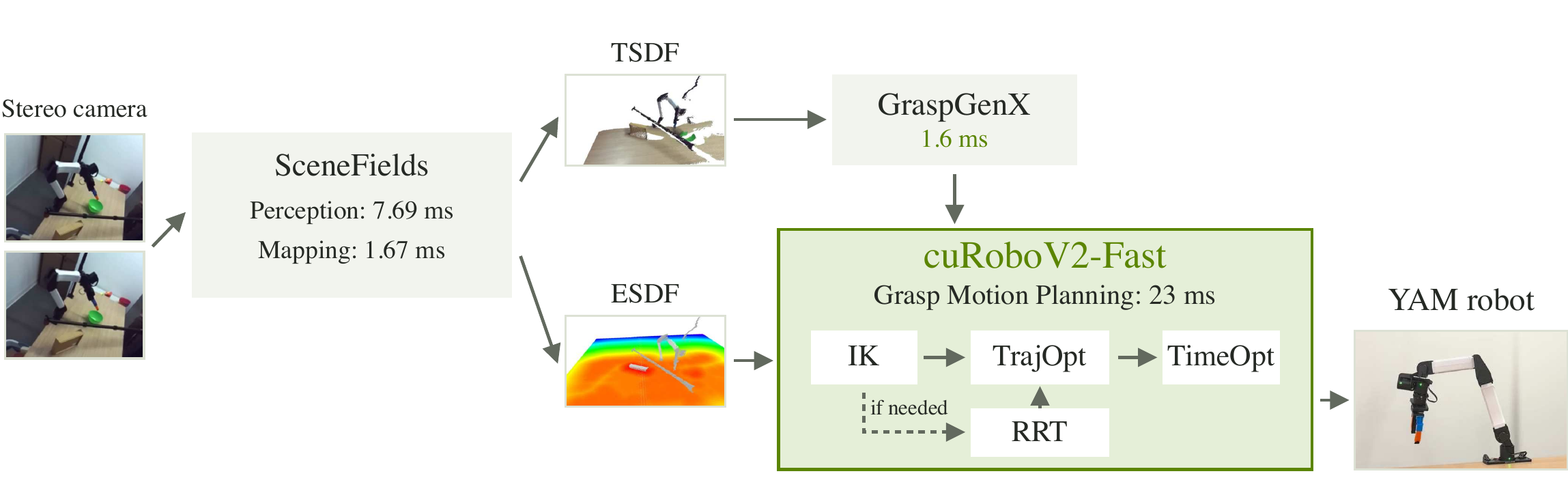}
\caption{Real-robot manipulation on one RTX 6000 Ada GPU.
SceneFields provides depth, detections, and robot masks; its maps
supply object surfaces to GraspGenX and collision distances to
cuRoboV2-Fast. Our planner selects a grasp and plans collision-free,
torque-limited pregrasp, approach, and lift motions in 23\,ms.
RRT initializes trajectory optimization when needed; transfer motions
are planned separately.
}
\label{fig:deployment_graph}
\end{figure*}

\subsection{Trajectory Optimization Computation}
\label{sec:trajopt_computation}
cuRoboV2-Fast reduces trajectory optimization time through faster
iterations and a smaller iteration budget
(Fig.~\ref{fig:trajopt_module_timings}). With inverse dynamics enabled,
each iteration is $2.8\times$ faster than cuRoboV2 on RTX 5090 and
$2.6\times$ faster on Jetson AGX Orin.

Part of this gain comes from avoiding repeated work during line search.
By reusing the cost at the current solution, cuRoboV2-Fast evaluates
972 robot states per iteration instead of cuRoboV2's 1,296.
The RTX 5090 module profiles show combined forward and backward inverse
dynamics taking 19.38\,$\mu$s versus 31.58\,$\mu$s for cuRoboV2
($1.6\times$ faster). Other costs and reductions are $4.2\times$ faster.
Combined scene and self-collision kernel time is 14.01\,$\mu$s for
cuRoboV2-Fast versus 16.22\,$\mu$s for cuRoboV2. Compared with cuRoboV1,
combined forward and backward kinematics are $2.9\times$ faster, while
L-BFGS and line search are $1.3\times$ faster.

The smaller iteration budget compounds these savings. Following the
convergence study in Section~\ref{sec:trajopt_convergence}, we use
75 iterations ($50 + 1\times25$) for cuRoboV2-Fast, compared with
300 for both baselines.
Together with the faster iterations, this reduces total time spent in
the optimization loop by a factor of $11.0$ on RTX 5090 and $10.5$ on
Jetson AGX Orin relative to cuRoboV2. The corresponding reductions
relative to cuRoboV1 are factors of $6.1$ and $3.0$, even though
cuRoboV1 omits inverse dynamics. These timings cover the optimization
iterations; initialization and other full-solve overhead are excluded.

\subsection{Simulated Planning Experiments}
We compare cuRoboV1~\cite{curobo_icra23}, cuRoboV2~\cite{curobo_v2}, and
cuRoboV2-Fast on a simulated YAM pick-and-place task that arranges five
cubes from one cluttered layout into a tray on RTX 5090.
Planning includes grasp selection, pregrasp, approach, lift, and transfer.
All three methods complete all five cubes, and the saved lift and transfer
trajectories pass independent held-object collision and payload torque audits.

Figure~\ref{fig:yam_tray_comparison} shows that cuRoboV2-Fast reduces both
planning time and planned motion duration.
Mean planning time per cube is 26.05\,ms, giving speedups of
$12.7\times$ over cuRoboV1 and $11.2\times$ over cuRoboV2.
Mean planned motion duration is 3.33\,s, which is 26.5\% shorter than
cuRoboV1 and 46.6\% shorter than cuRoboV2.
Together, these gains reduce mean cycle time, measured as planning plus
planned motion, to 3.36\,s per cube: 31.0\% shorter than cuRoboV1 and
48.6\% shorter than cuRoboV2.

In a separate simulated YAM transfer between fixed pickup and placement
locations, we vary the payload from 100\,g to 800\,g
(Fig.~\ref{fig:yam_payload_comparison}). The heavier payload changes the
optimized motion: movement shifts from joint 2 to joint 5 to respect
joint 2's torque limit, illustrating how payload-aware planning adapts
the path to the load.

\subsection{Real-Robot Deployment}
\begin{figure}[!htbp]
\centering
\includegraphics[width=\columnwidth]{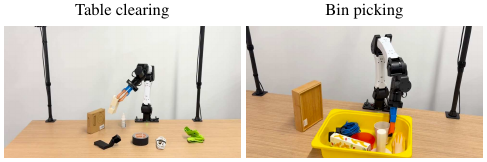}
\par\vspace{3pt}
\includegraphics[width=\columnwidth]{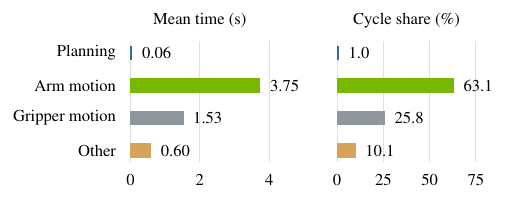}
\caption{Real YAM table clearing and bin picking with cuRoboV2-Fast.
Mean cycle breakdown over 53 completed sequences; planning accounts for
only 1.0\%, including failed planning calls and retries within each cycle.
Other includes residual processing and waits.}
\label{fig:real_robot_pick_place}
\label{fig:deployment_timing_main}
\end{figure}

We deploy cuRoboV2-Fast on the I2RT YAM for tabletop clearing and bin
picking (Fig.~\ref{fig:real_robot_pick_place}).
One RTX 6000 Ada runs SceneFields~\cite{scenefields2026},
GraspGenX~\cite{graspgenx2026}, and our planner
(Fig.~\ref{fig:deployment_graph}). SceneFields provides depth, detections,
and robot masks. It builds a TSDF with 5\,mm voxels
for object surfaces used by GraspGenX and an ESDF with 20\,mm voxels for
collision checking by our planner.

\begin{figure}[!htbp]
\centering
\includegraphics[width=\columnwidth]{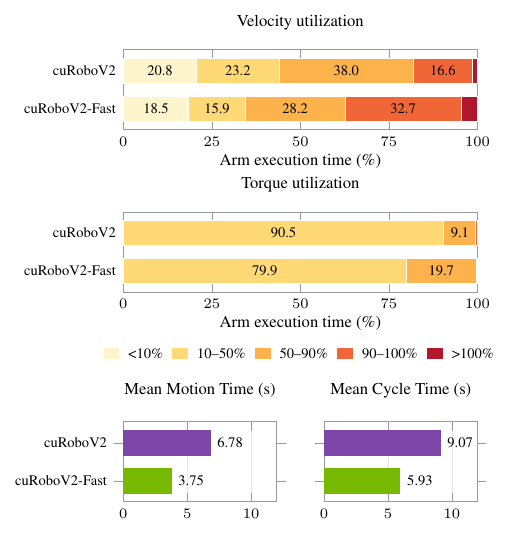}
\caption{Arm utilization, motion time, and cycle time for real-robot object
transfers. Utilization uses the most utilized joint at each instant.
cuRoboV2-Fast nearly doubles time at 90--100\% velocity utilization and
reduces motion and cycle times by 44.8\% and 34.6\%, respectively.
}
\label{fig:deployment_joint_limits}
\label{fig:deployment_cycle_time}
\end{figure}

Measured joint feedback shows that cuRoboV2-Fast uses more of the available
velocity range during arm execution (Fig.~\ref{fig:deployment_joint_limits}).
Relative to cuRoboV2, the fraction of motion time at 90--100\% velocity
utilization nearly doubles, while the fraction below 50\% utilization
decreases by 21.7\%. Torque utilization remains below 90\% for 99.6\% of
motion time with cuRoboV2-Fast, and all recorded torques remain within
limits for both methods. Executed velocities can exceed limits despite
velocity-feasible plans, highlighting the importance of considering
torque constraints alongside velocity constraints when planning motions
for execution.

Spending a larger fraction of motion time at higher velocities translates
into shorter arm execution times. Across approximately 50 completed
object-transfer cycles for cuRoboV2-Fast and 15 for cuRoboV2, mean arm
execution time decreases by 44.8\%. This reduction leads to a 34.6\%
decrease in mean cycle time (Fig.~\ref{fig:deployment_cycle_time}).

We break down the time in a cycle across motion planning, arm execution,
gripper execution, and other processing and waits
(Fig.~\ref{fig:deployment_timing_main}). For cuRoboV2-Fast, these average
0.06, 3.75, 1.53, and 0.60\,s, respectively. Arm execution
accounts for 63.1\% of cycle time, while planning accounts for only 1.0\%.

\section{Discussion}
\label{sec:discussion}

The planning-time improvements come from reducing both the number of
optimization iterations and the cost of each iteration. The convergence
experiment shows that our B-spline formulation reaches comparable success
with fewer initial iterations in the evaluated configuration, while
accelerated robot computations reduce the cost of refinement with inverse
dynamics. Together with fast route initialization, these improvements yield
full-pipeline speedups while maintaining similar success to cuRoboV2 across
payloads. The gains also extend to Jetson AGX Orin, where cuRoboV2-Fast plans
faster than cuRoboV1 + TOPPRA on RTX 5090, supporting deployment on
embedded hardware.

Efficient execution requires optimizing both the path and its timing.
The payload benchmarks show that shorter joint-space paths do not
necessarily produce shorter motions under actuator constraints.
Payload-aware trajectory optimization can adapt the path to the load,
while TimeOpt redistributes traversal time along the resulting path.
Its 15.7\% reduction in mean motion duration beyond uniform timestep
refinement shows that timing remains an opportunity for improvement after
geometry optimization. Local spline support yields banded Newton systems
and parallel constraint evaluation, making this additional optimization
inexpensive on the GPU. These results support TimeOpt as a practical way
to improve execution time without requiring globally optimal timing.

The real-robot experiments connect these improvements to execution:
cuRoboV2-Fast spends a larger fraction of arm-motion time near the velocity
limits and reduces mean arm execution time by 44.8\% relative to
cuRoboV2. Shorter arm motions also reduce completed cycle time, although
gripper execution and other processing remain substantial contributors.
Planning accounts for only 1.0\% of the measured cycle time, so further
planning-latency reductions alone would have limited impact on these
cycles. Improving the motions produced by the planner therefore remains
central to faster manipulation, alongside integration with perception,
grasp generation, and execution.

\subsection{Limitations}
\label{sec:limitations}

Our approach addresses motion planning holistically, enabling high success
in cluttered scenes while reducing both computation time and motion duration.
Although it does not guarantee global optimality, it achieves a mean
pick-and-transfer cycle time of 5.9\,s on a real robot. Planning accounts for
only 1.0\% of cycle time in this deployment and is no longer a bottleneck.
The main remaining bottleneck lies in contact-rich manipulation, particularly
achieving reliable grasps and successful object interactions.

\FloatBarrier
\IEEEtriggeratref{23}
\bibliographystyle{IEEEtran}
\bibliography{references}

\end{document}